\documentclass[journal]{IEEEtran}
\usepackage{graphicx}
\usepackage{subfigure}
\graphicspath{{pic/}}
\usepackage{cite}
\usepackage{amsthm}
\usepackage{amsmath}

\begin{document}

\title{LRA: an accelerated rough set framework based on local redundancy of attribute for feature selection}

\author{Shuyin~Xia, 
	Wenhua~Li, 
	Guoyin~Wang*, 
	Xinbo~Gao, 
	Changqing~Zhang, 
	Elisabeth~Giem
	\thanks{S. Xia, W. Li, G. Wang \& X. Gao are with the Chongqing Key Laboratory of Computational Intelligence, Chongqing University of Telecommunications and Posts, Chongqing, 400065, China. E-mail: xiasy@cqupt.edu.cn, 1025476698@qq.com, wanggy@cqupt.edu.cn, gaoxb@cqupt.edu.cn}
	
	\thanks{C. Zhang is with the College of Intelligence and Computing, Tianjin University, 300072, China.
		E-mail:zhangchangqing@tju.edu.cn}
	
	\thanks{E. Giem is with the Department of Computer Science and Engineering, University of California Riverside, Riverside, CA, 92521. E-mail: gieme01@ucr.edu}
}

\maketitle

\begin{abstract}
In this paper, we propose and prove the theorem regarding the stability of attributes in a decision system. Based on the theorem, we propose the LRA framework for accelerating rough set algorithms. It is a general-purpose framework which can be applied to almost all rough set methods significantly . Theoretical analysis guarantees high efficiency. Note that the enhancement of efficiency will not lead to any decrease of the classification accuracy. Besides, we provide a simpler prove for the positive approximation acceleration framework.
\end{abstract}


\section{Introduction\label{sec:introduction}}

\IEEEPARstart{R}{ough} set (RS) theory is a theory that mathematically analyzes the expression of incomplete data, imprecise knowledge, and learning induction. Introduced by Polish mathematician Zdzis\l aw Pawlak in 1982 \cite{Zdzisaw1982Rough}, the foundational concepts are information granulation and approximation. An equivalence relation granulates discrete samples into disjoint granules of equivalent sample information, and the uncertainty of the information granules is characterized by upper and lower approximation bounds to obtain an approximation of the arbitrary knowledge in the sample. Rough set theory does not require any prior knowledge to handle imprecise or uncertain problems; it has strong objectivity. Rough set theory has been extended to the fuzzy rough set model \cite{lin2018attribute,wang2019fuzzy,radzikowska2002comparative}, the probabilistic rough set model \cite{Yao2011The,Ziarko2005Probabilistic,Fang2016Probabilistic}, the covering rough set model \cite{Zhu2006Relationships,Han2018Covering}, the neighborhood rough set model \cite{Qing2008Numerical,Hu2008Neighborhood,Changzhong2018Attribute}, multigranular rough set model \cite{raghavan2011some,tripathy2013approximate,tripathy2018properties}, and more. In recent years, rough set theory has become a subject of ever-increasing interest and enthusiasm among academics. Many knowledge discovery systems in rough set theory are being explored, which has resulted in the successful application of rough set theory to machine learning, decision analysis, process control, pattern recognition, data mining, and other fields \cite{Aggarwal2017Rough,Gediga2001Rough,Saltos2017Dynamic,guan1998rough,D1998Uncertainty,Tsang2016Feature,rehman2018sdmgrs,Herawan2010A,yao2014web}.

\section{A novel framework to accelerate rough sets\label{sec:novel}}

We propose and prove two theorems, which we call the \textit{stability of redundancy attribute} (SR theorem) and \textit{stability of local redundancy attribute} (SLR theorem), that can be generally used to accelerate almost all existing rough set algorithms.

\textbf{\emph{Theorem 1}} {\bf{\{The stability of redundancy (SR) attribute\}.}} Given a decision system $ \left \langle U,C,D \right \rangle$, let $ R\subseteq C $ and $ a\in C $, $ b\in C $ are given attributes. If $ U'/R = U'/(R + b) $, where $ U'=U-POS_{R}(D) $, $b$ is a redundant attribute relative to $ R + a $.

\textit{Proof:} Let $ U'/R = \{X_{1},X_{2},...,X_{l} \} $, $ U'/b = \{X_{1}^{'},X_{2}^{'},...,X_{l'}^{'} \} $, $ U'/a = \{X_{1}^{''},X_{2}^{''},...,X_{l''}^{''} \} $, we have,
\begin{equation}
	X_{1}\cup X_{2}\cup ...\cup X_{l} = U' , X_{i}\cap X_{j} = \O ( i\neq j ),
\end{equation}
\begin{equation}
	X_{1}^{'}\cup X_{2}^{'}\cup ...\cup X_{l'}^{'} = U' ,  X_{i}^{'}\cap X_{j}^{'} = \O ( i\neq j ),
\end{equation}
\begin{equation}
	X_{1}^{''}\cup X_{2}^{''}\cup ...\cup X_{l''}^{''} = U' ,  X_{i}^{''}\cap X_{j}^{''} = \O ( i\neq j ).
\end{equation}
\begin{equation}
	U'/R = U'/(R + b),
\end{equation}
\begin{equation}
	\begin{aligned}
		\{ X_{1},X_{2},...,X_{l} \} =  \{X_{1}\cap X_{1}^{'}, X_{1}\cap X_{2}^{'},..., X_{2}\cap X_{1}^{'}, \\ 
		X_{2}\cap X_{2}^{'},..., X_{i}\cap X_{j}^{'},..., X_{l}\cap X_{l'}^{'} \},
	\end{aligned}
\end{equation}
\begin{equation}
	X_{i} \subseteq X_{j}^{'}  ( i = 1,2,...,l ,  j=1,2,...,l' ).
\end{equation}
\begin{equation}
	\begin{aligned}
		U'/(R+a) = \{X_{1}\cap X_{1}^{''}, X_{1}\cap X_{2}^{''},..., X_{2}\cap X_{1}^{''}, \\
		X_{2}\cap X_{2}^{''},..., X_{i}\cap X_{j}^{''},..., X_{l}\cap X_{l''}^{''} \}.
	\end{aligned}
\end{equation}

Since $ X_{i}\cap X_{j}^{''} \subseteq X_{i} $, combined with formula (6),
\begin{equation}
	X_{i}\cap X_{j}^{''} \subseteq X_{j}^{'} ,
\end{equation}
\begin{equation}
	X_{i}\cap X_{j}^{''}\cap X_{j}^{'} = X_{i}\cap X_{j}^{''} .
\end{equation}
\begin{equation}
	\begin{aligned}
		U'/(R+a+b) = \{X_{1}\cap X_{1}^{''}\cap X_{1}^{'}, X_{1}\cap X_{2}^{''}\cap X_{1}^{'},\\ 
		..., X_{2}\cap X_{1}^{''}\cap X_{1}^{'}, X_{2}\cap X_{2}^{''}\cap X_{1}^{'},...,\\
		X_{i}\cap X_{j}^{''}\cap X_{j}^{'},..., X_{l}\cap X_{l''}^{''}\cap X_{l'}^{'} \}.
	\end{aligned}
\end{equation}

Combined with formula (9),
\begin{equation}
	\begin{aligned}
		(10)=\{X_{1}\cap X_{1}^{''}, X_{1}\cap X_{2}^{''},..., X_{2}\cap X_{1}^{''},\\
		X_{2}\cap X_{2}^{''},..., X_{i}\cap X_{j}^{''},..., X_{l}\cap X_{l''}^{''} \}=(7),
	\end{aligned}
\end{equation}
\begin{equation}
	U'/(R+a+b) = U'/(R+a),
\end{equation}
\begin{equation}
	POS_{R + a + b}(D) = POS_{R + a}(D).
\end{equation}

$ \Rightarrow $ $b$ also is a redundant attribute relative to $ R + a $.\qed

That is, the redundant attribute relative to the child attribute set is also the redundant attribute relative to its parent attribute set. In other words, the redundancy of the attribute is stable.

Inspired by Theorem 1, we described the definition of the active region and the non-active region.

\textbf{\emph{Definition 1.}} {\bf{\{Active Region and Non-Active Region\}.}} Let $ \left \langle U,C,D \right \rangle $ be a decision system. Let $ R\subseteq C $  and $ a\in C $ (but $ a\notin R $) be a given attribute. The equivalence class that $ U $ is divided into under the attribute set $ R $ is $ U/R = \left \{X_{1},X_{2},...,X_{i},X_{i+1},...,X_{l} \right \} $ and the equivalence class that $ U $ is divided into under the attribute $ a $ is $ U/a = \left \{X_{1}^{'},X_{2}^{'},...,X_{k}^{'},X_{k+1}^{'},...,X_{s}^{'} \right \} $. For $\forall$  $j\in{1,2...,i}$, if $ X_{j} \subseteq X_{t}^{'} $($ t=1,2,...,s $) exists, then we define set $ X_{1}\cup X_{2}\cup ...\cup X_{i} $ as the non-active region of attribute $ a $ and set $ X_{i+1}\cup ...\cup X_{l} $ as the active region of the attribute $ a $.

\textbf{\emph{Theorem 2}} {\bf{\{The stability of local redundancy (SLR) attribute\}.}} Given a decision system $ \left \langle U,C,D \right \rangle$, let $ R\subseteq C $ and $ a\in C $ (but $ a\notin R $) be a given attribute. The equivalence class that $ U $ is divided into under the attribute set $ R $ is $ U/R = \left \{X_{1},X_{2},...,X_{i},X_{i+1},...,X_{l} \right \} $ and the equivalence class that $ U $ is divided into under the attribute $ a $ is $ U/a = \left \{X_{1}^{'},X_{2}^{'},...,X_{k}^{'},X_{k+1}^{'},...,X_{s}^{'} \right \} $. Let the active region of $ a $ be $ U_{a} = X_{i+1}\cup ...\cup X_{l} $, we only need to pay attention to the active region of $ a $ to determine whether $ a $ is a non-redundant attribute relative to the attribute set $ R $.

\textit{Proof:} let $ U/R = \left \{X_{1},X_{2},...,X_{i},X_{i+1},...,X_{l} \right \} $, $ U_{a} = X_{i+1}\cup ...\cup X_{l} $ is the active region of $ a $, and $ U_{a}^{'} = X_{1}\cup X_{2}\cup ...\cup X_{i} $ is the non-active region of $ a $.

Since $ U_{a}^{'} $ is the non-active region of $ a $, we have
\begin{equation}
	X_{j} \subseteq X_{t}^{'} ( j = 1,2,...,i ,  t=1,2,...,s ),
\end{equation}
\begin{equation}
	X_{j} \cap X_{t}^{'} = X_{j}.
\end{equation}
\begin{equation}
	\begin{aligned}
		U/(R + a) = \{X_{1}\cap X_{1}^{'}, X_{1}\cap X_{2}^{'},..., X_{2}\cap X_{1}^{'}, \\
		X_{2}\cap X_{2}^{'},..., X_{i}\cap X_{1}^{'},..., X_{l}\cap X_{s}^{'} \} \\
		= \left \{X_{1},X_{2},...,X_{i},X_{i+1}\cap X_{1}^{'},X_{i+1}\cap X_{2}^{'},..., X_{l}\cap X_{s}^{'} \right \}\\
		= \left \{U_{a}^{'},X_{i+1}\cap X_{1}^{'},X_{i+1}\cap X_{2}^{'},..., X_{l}\cap X_{s}^{'} \right \}
	\end{aligned}
\end{equation}

Focusing only on the active region of $ a $, we have,

$ U/(R+a) = \{U_{a}^{'}, X_{i+1}\cap X_{1}^{'},X_{i+1}\cap X_{2}^{'},..., X_{l}\cap X_{s}^{'} \} $ is the same as the formula above.

$ \Rightarrow $ Focusing only on the active region of $ a $ can determine the redundancy of $ a $ relative to the attribute set $ R $.\qed

\section{Experiments\label{sec:experiments}}

In this section, we implement our framework on the neighborhood rough set method and compare it with both the classic rough set method and the fast NRS method FARNeMF, or Forward Attribute Reduction Based on Neighborhood Rough Sets and Fast Search, as our baselines. 

We demonstrate the effectiveness of the proposed LRA framework on selected UCR benchmark data sets (http://archive.ics.uci.edu/ml/datasets.html), which are described in detail in Table \ref{tab:table_1}.

\begin{table}[htp]
	\renewcommand{\arraystretch}{1.3}
	\caption{Dataset Information}
	\label{tab:table_1}
	\centering	
	\begin{tabular}{clcccc}
		\hline
		{}&{Dataset} &{Samples} & {\begin{tabular}[c]{@{}c@{}}Numerical \\ Features\end{tabular}} & {\begin{tabular}[c]{@{}c@{}}Categorical \\ Features\end{tabular}} & {Class} \\ 
		\hline
		1                        & anneal                       & 798                          & 6                                       & 32                                        & 5                          \\
		2                        & credit                       & 690                          & 6                                       & 9                                         & 2                          \\
		3                        & german                       & 1000                         & 7                                       & 12                                        & 2                          \\
		4                        & heart1                       & 270                          & 7                                       & 6                                         & 2                          \\
		5                        & hepatitis                    & 155                          & 6                                       & 13                                        & 2                          \\
		6                        & horse                        & 368                          & 7                                       & 16                                        & 2                          \\
		7                        & iono                         & 351                          & 34                                      & 0                                         & 2                          \\
		8                        & wdbc                         & 569                          & 30                                      & 0                                         & 2                          \\
		9                       & zoo                         & 101                          & 0                                      & 16                                         & 7                          \\
		10                       & mocap                          & 78000                          & 33                                       & 0                                        & 2                          \\
		\hline
	\end{tabular}
\end{table}

\begin{figure}[htp]
	\centering
	\subfigure[]{\includegraphics[width=0.24\textwidth]{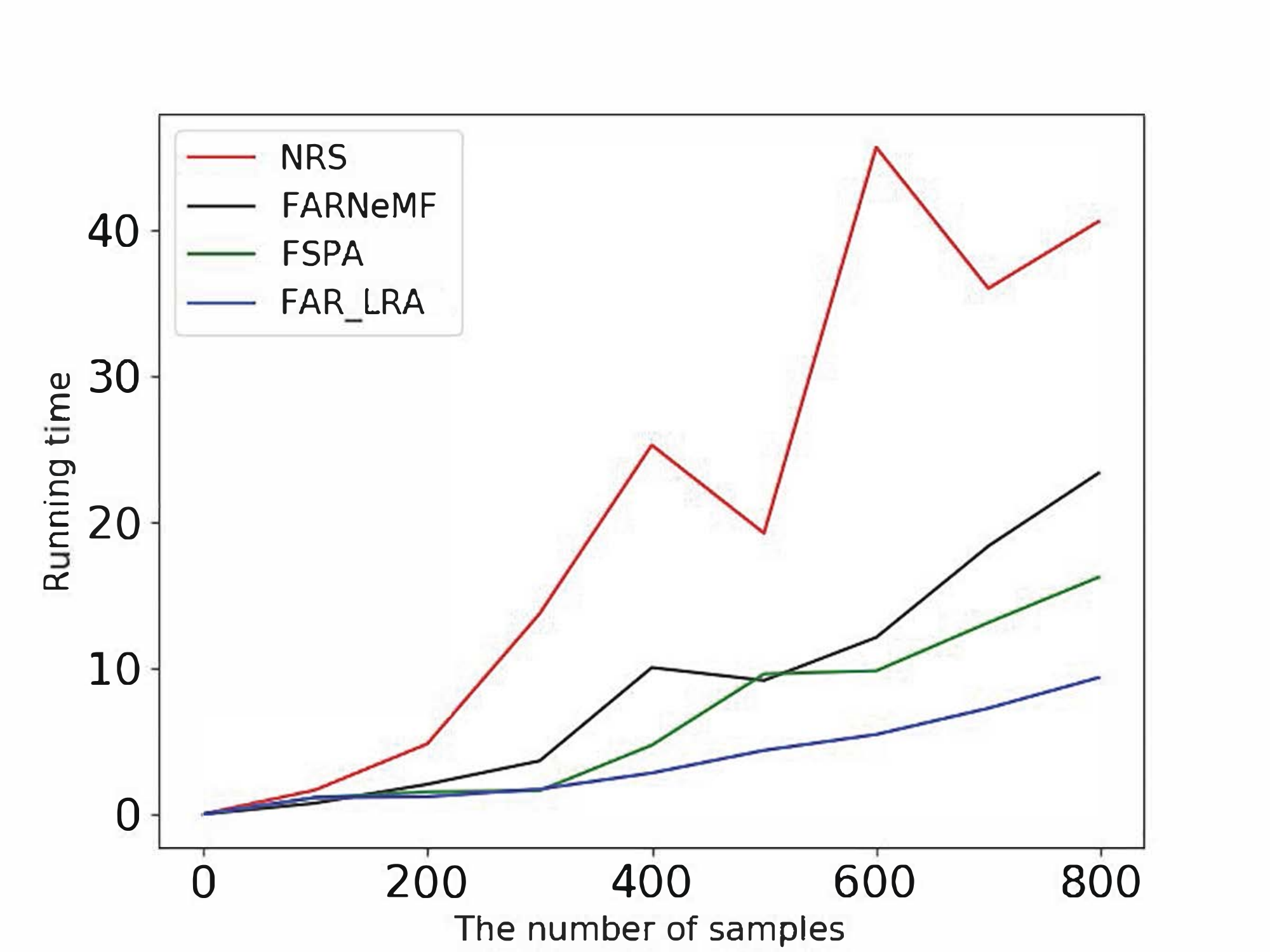}}
	\subfigure[]{\includegraphics[width=0.24\textwidth]{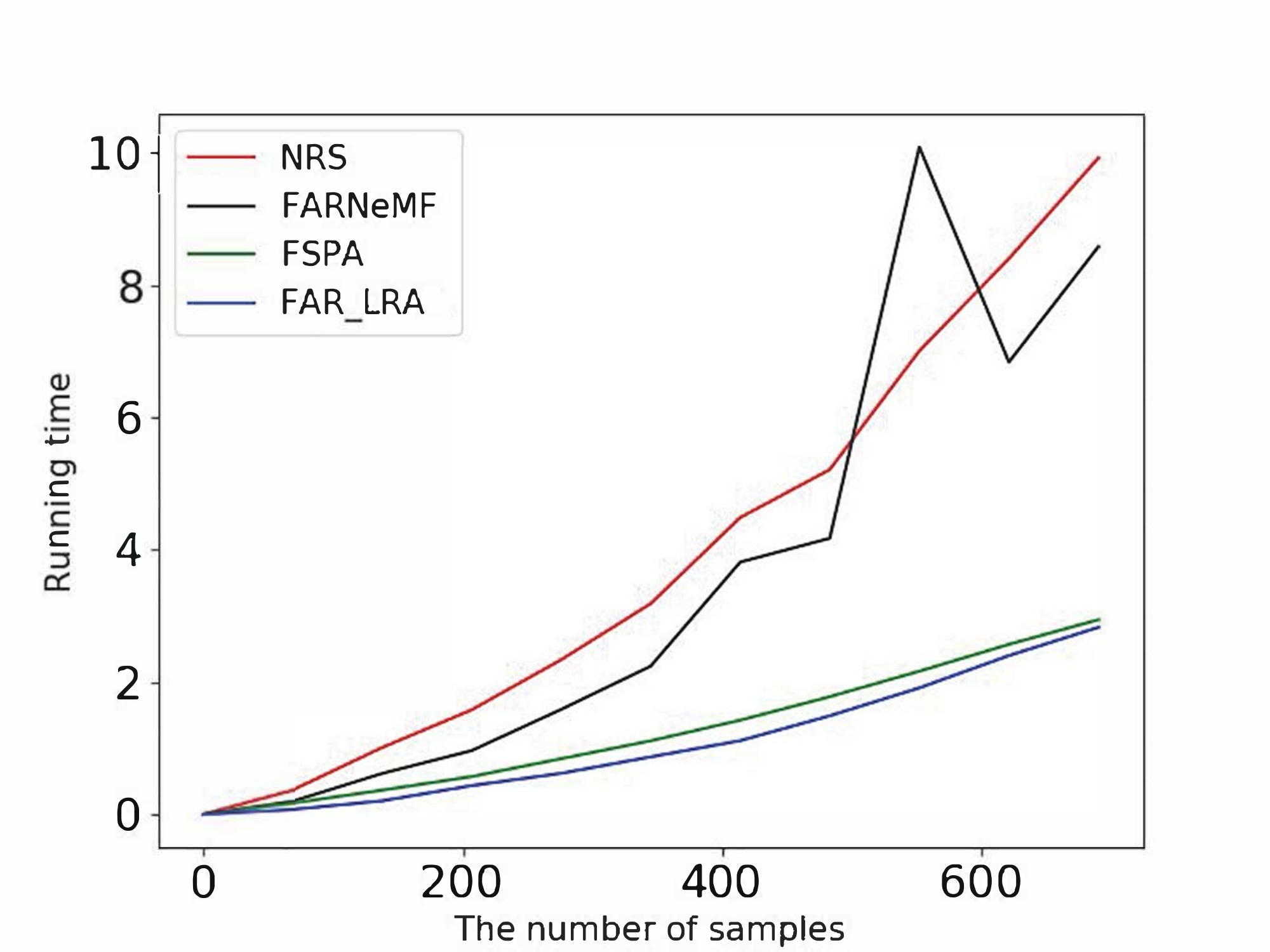}}
	\subfigure[]{\includegraphics[width=0.24\textwidth]{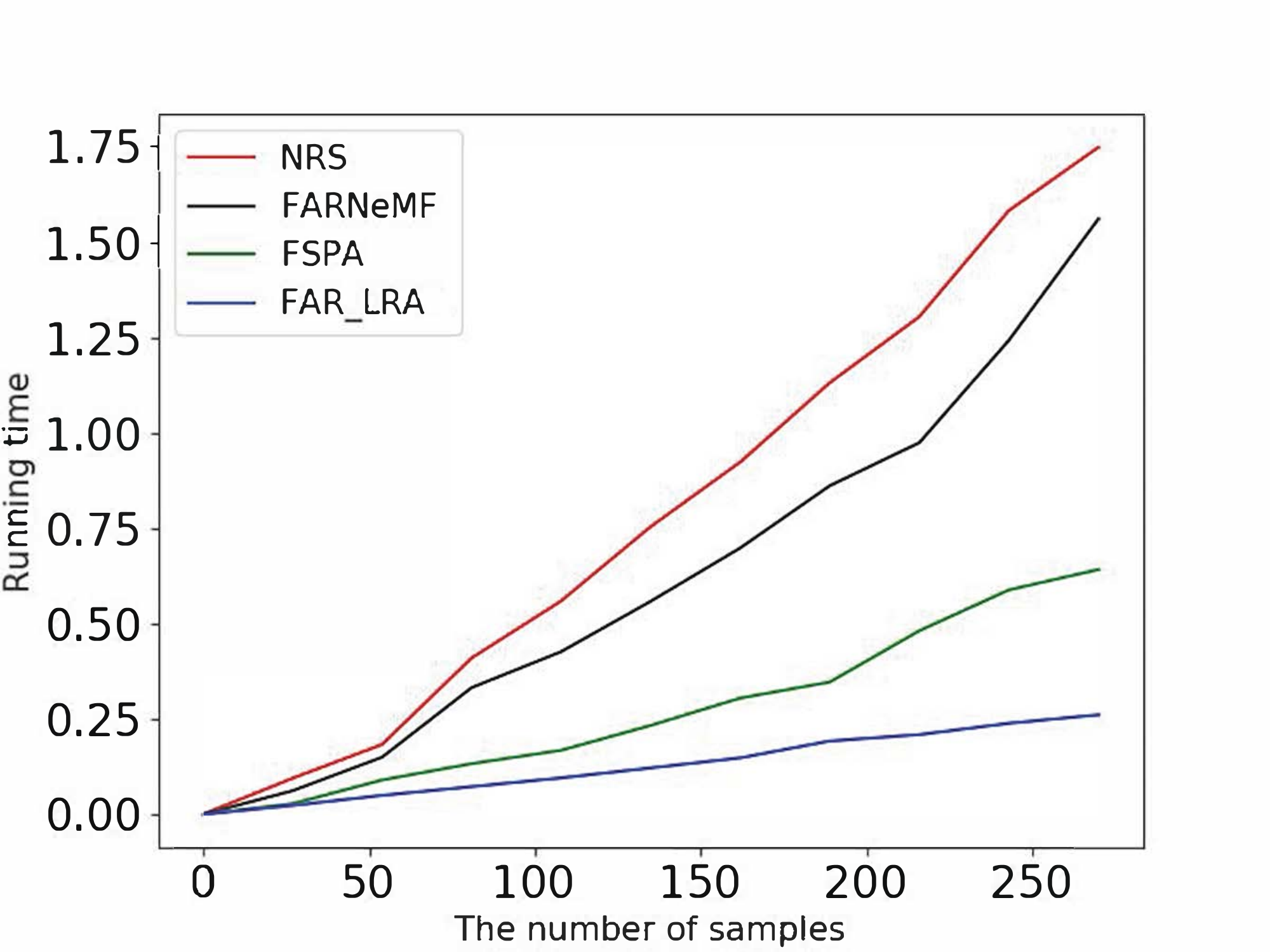}}
	\subfigure[]{\includegraphics[width=0.24\textwidth]{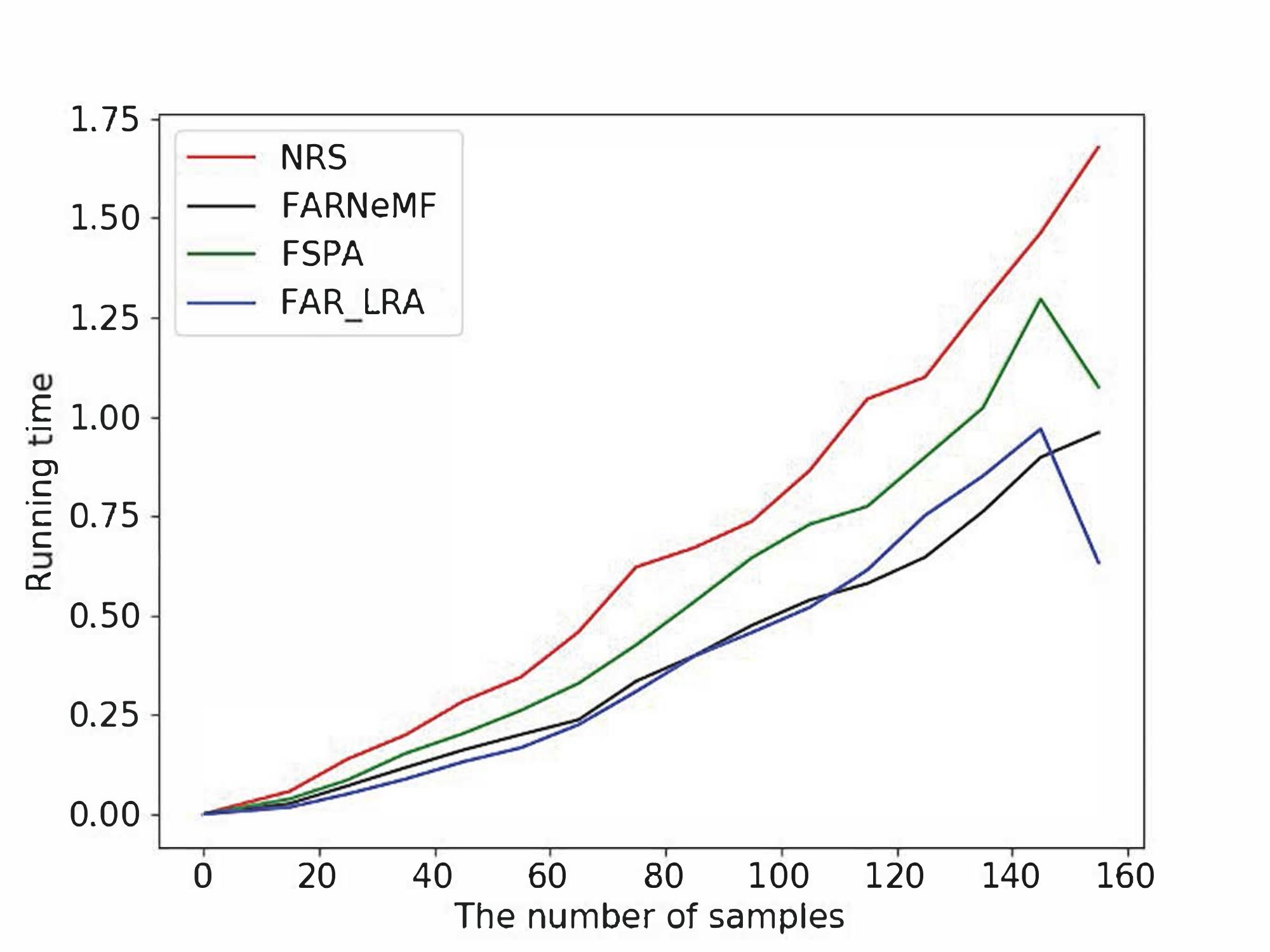}}
	\subfigure[]{\includegraphics[width=0.24\textwidth]{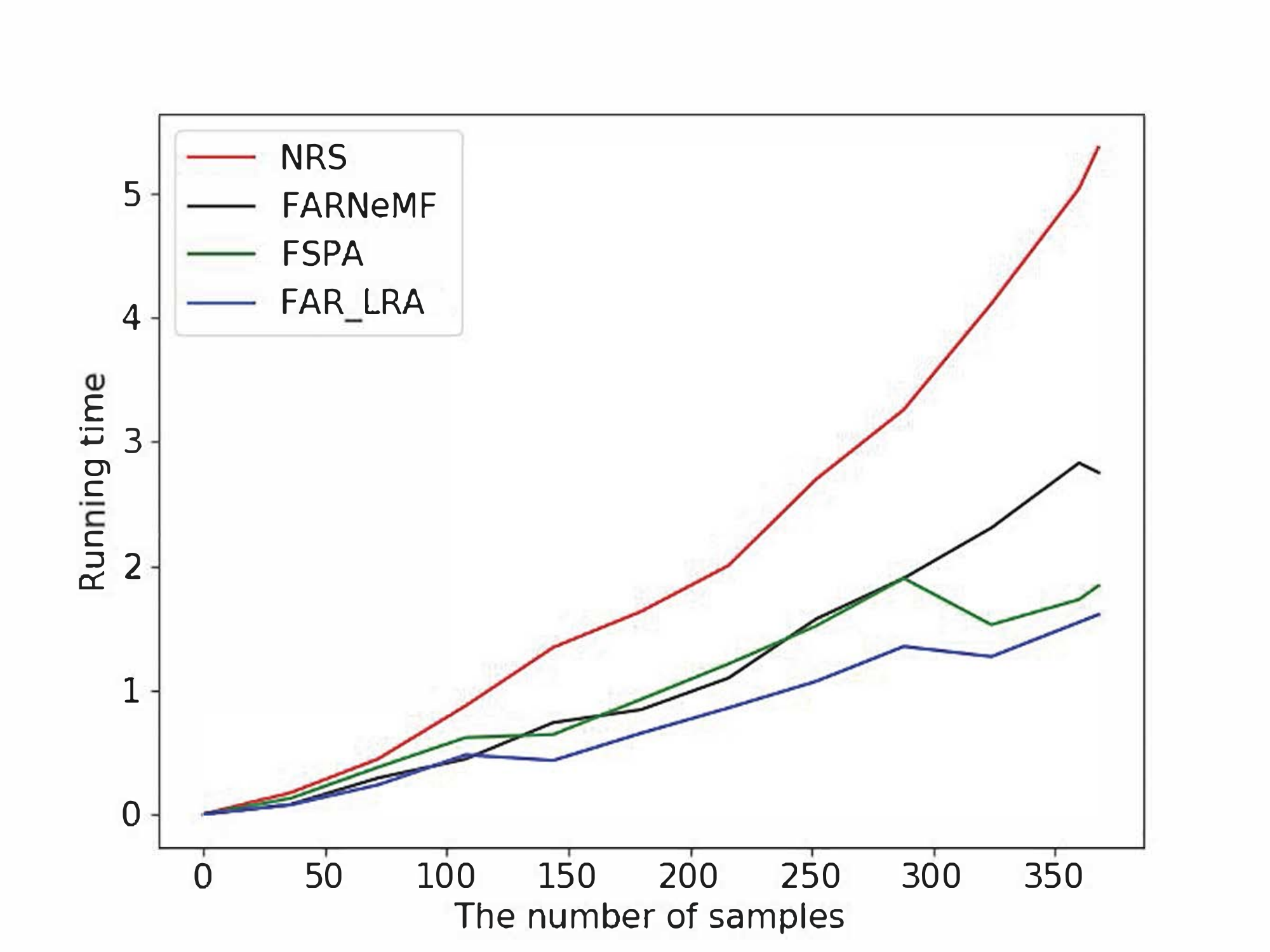}}
	\subfigure[]{\includegraphics[width=0.24\textwidth]{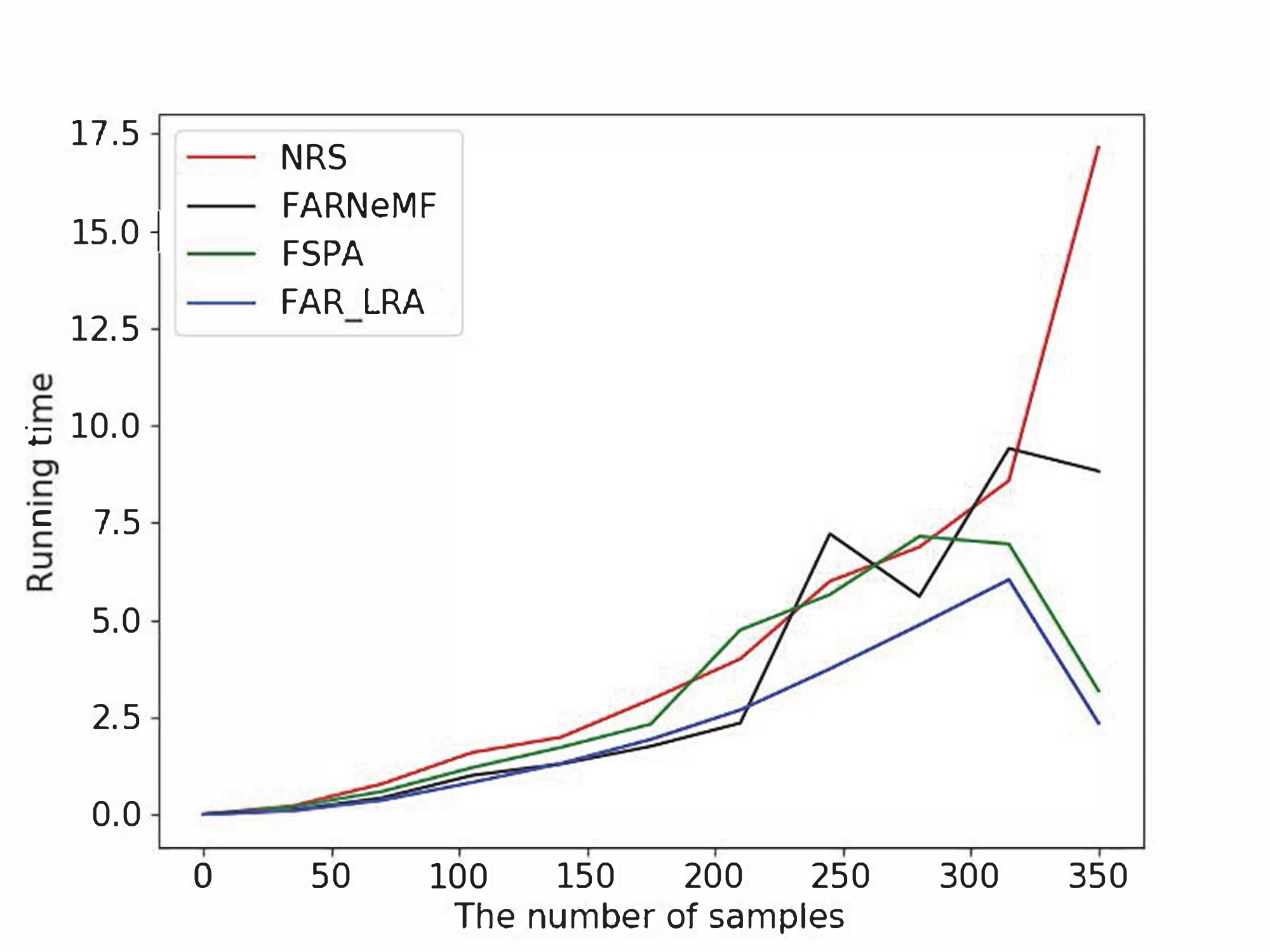}}
	\subfigure[]{\includegraphics[width=0.24\textwidth]{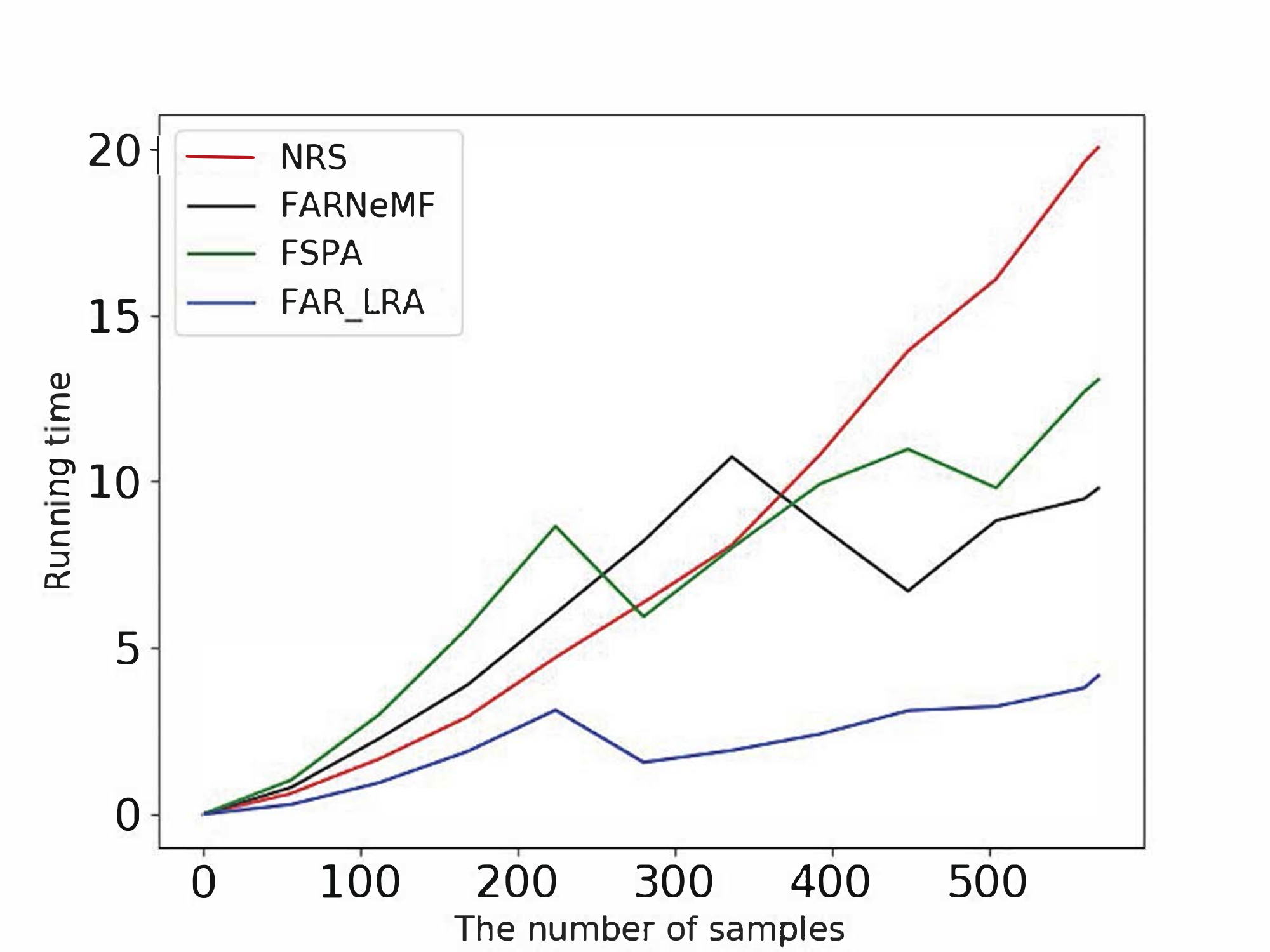}}
	\subfigure[]{\includegraphics[width=0.24\textwidth]{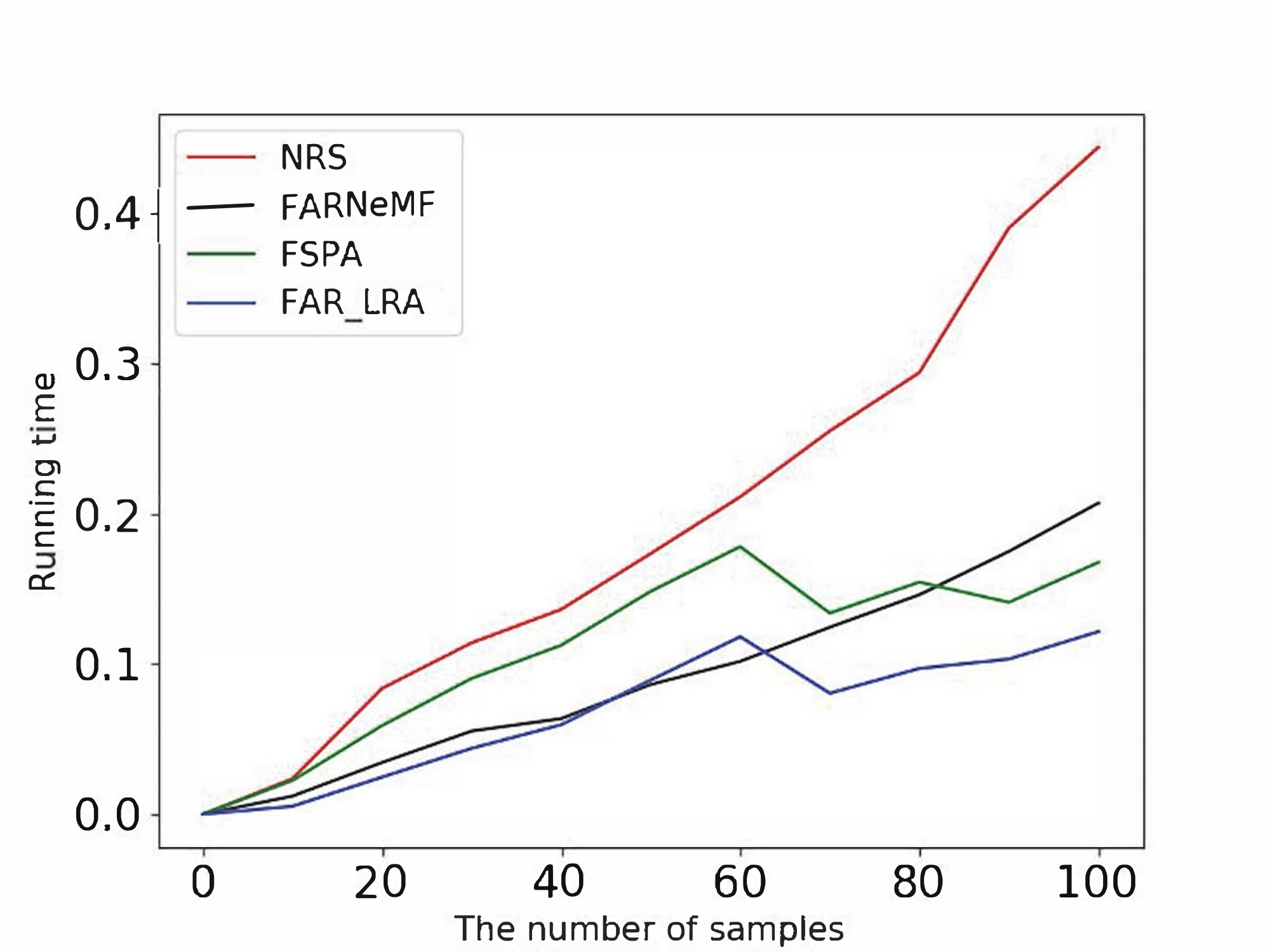}}
	\subfigure[]{\includegraphics[width=0.24\textwidth]{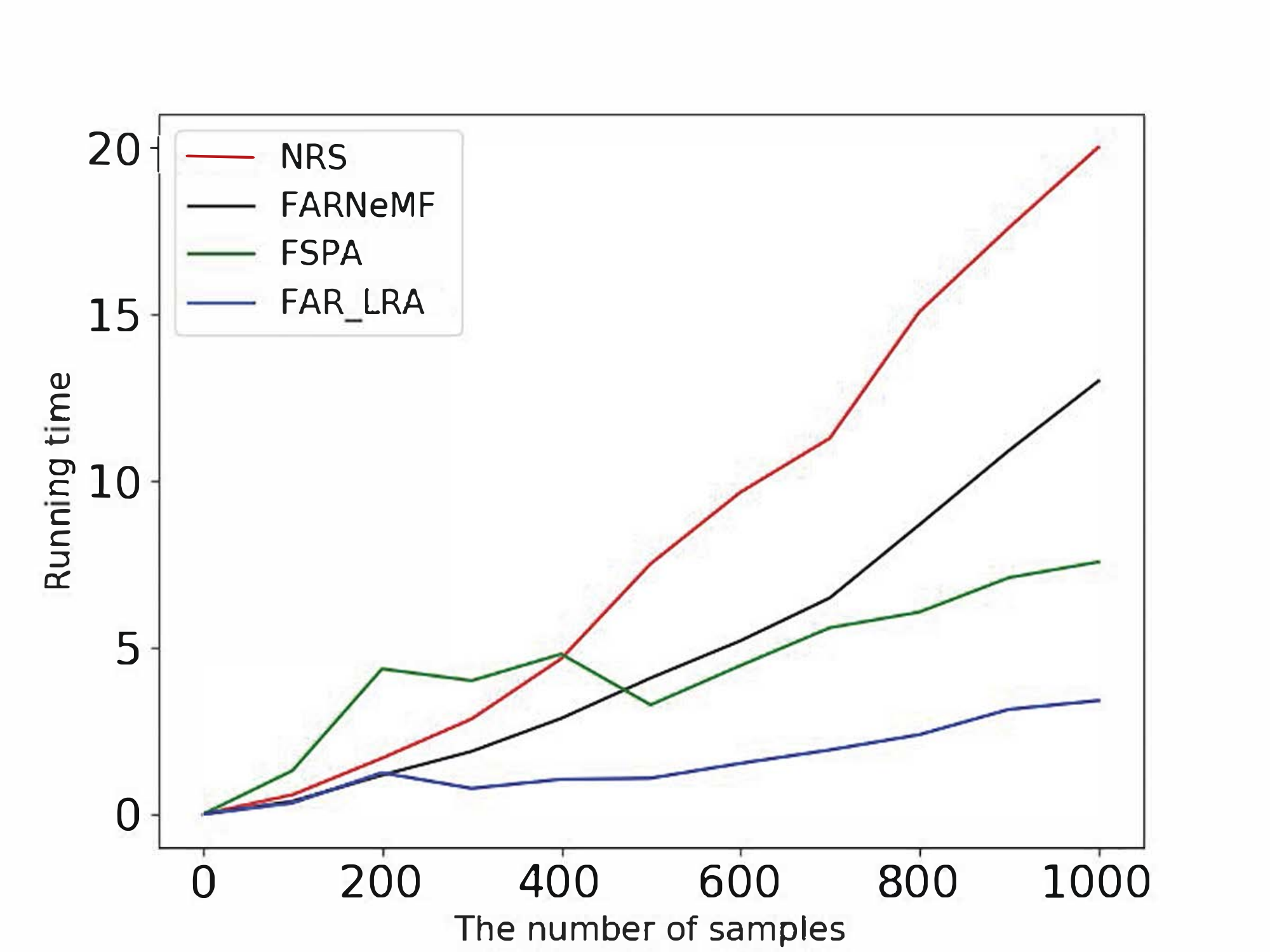}}
	\subfigure[]{\includegraphics[width=0.24\textwidth]{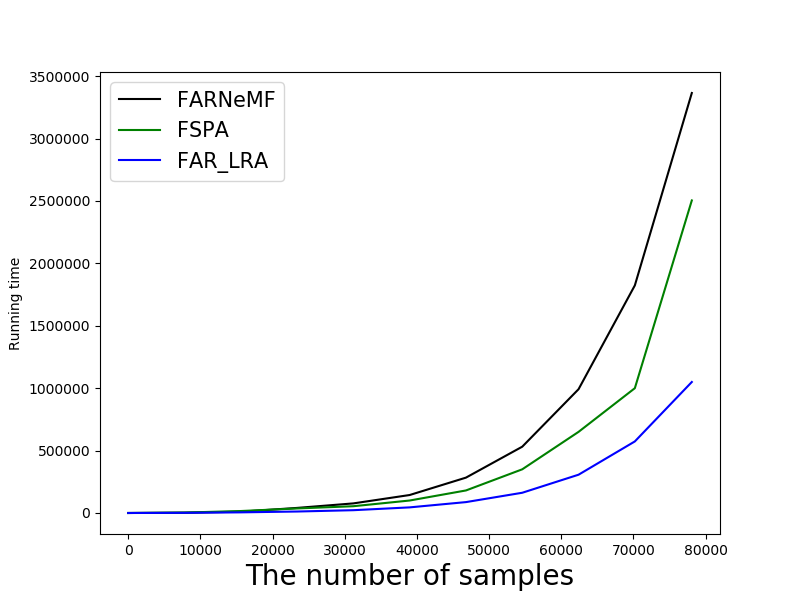}}
	\caption{Efficiency Comparison on the Following Datasets: (a) anneal (b) credit (c) heart1 (d) hepatitis (e) horse (f) iono (g) wdbc (h) zoo (i) german (j) mocap. The unit of the ordinate axis is seconds.}
	\label{fig:Fig3}
\end{figure}

Figure \ref{fig:Fig3} is a comparative experiment result of the neighborhood rough set. The red solid line represents the neighborhood rough set algorithm (NRS), the black solid line represents the FARNeMF method accelerated only by Theorem 1, the green solid line represents the method of FSPA applied to the neighborhood rough set, and the blue solid line represents the version (FAR$\_$LRA) accelerated by three theorems regarding the stability of attributes. We fixed the neighborhood radius at 0.16 to eliminate the neighborhood radius as a variable in the experiment and compare only the efficiency between the algorithms.

We can see that the blue solid curve is below the other curves in most cases, and the LRA framework allows for a much more efficient algorithm than the baseline comparison algorithm whether it is applied to classic rough sets or neighborhood rough sets. This is because the LRA framework can significantly reduce the number of objects and attributes considered in the iterative process, thus reducing the computation times of the positive region of joint attributes. In addition, as the number of samples increases, the advantage of LRA based algorithms becomes more large.

\section{Conclusion\label{Conclusion}}

This paper presents and proves two theorems regarding the stability of attributes in a decision system. Based on two theorems, we propose the LRA framework for accelerating rough set algorithms. Theoretical analysis guarantees its high efficiency. Our experimental results also demonstrate that our LRA framework can considerably accelerate these rough set algorithms in most cases ranging from several times faster up to ten times faster. In addition, the neighborhood rough set method accelerated with LRA overcomes a shortcoming of existing neighborhood rough set methods, namely that it still functions when the neighborhood radius is large where existing neighborhood rough set methods will fail. In spite of these advances, there are still some interesting issues that it will be valuable to investigate in the future. To improve its performance in efficiency in large-scale data, we will consider parallelization.

\end{document}